\documentclass[runningheads]{llncs}
\usepackage{amsfonts}
\usepackage[utf8]{inputenc}

\title{On the independence between phenomenal consciousness and computational intelligence}
\titlerunning{On the independence between consciousness and intelligence}
\author{Eduardo C. Garrido Merchán, Sara Lumbreras}

\institute{Universidad Pontificia de Comillas, Madrid, Spain
\email{ecgarrido@icade.comillas.edu} \and
Universidad Pontificia de Comillas, Madrid, Spain
\email{slumbreras@comillas.edu}}

\begin{document}

\maketitle

\abstract{Consciousness and intelligence are properties commonly understood as dependent by folk psychology and society in general. The term artificial intelligence and the kind of problems that it managed to solve in the recent years has been shown as an argument to establish that machines experience some sort of consciousness. Following Russell's analogy, if a machine is able to do what a conscious human being does, the likelihood that the machine is conscious increases. However, the social implications of this analogy are catastrophic. Concretely, if rights are given to entities that can solve the kind of problems that a neurotypical person can, does the machine have potentially more rights that a person that has a disability? For example, the autistic syndrome disorder spectrum can make a person unable to solve the kind of problems that a machine solves. We believe that the obvious answer is no, as problem solving does not imply consciousness. Consequently, we will argue in this paper how phenomenal consciousness and, at least, computational intelligence are independent and why machines do not possess phenomenal consciousness, although they can potentially develop a higher computational intelligence that human beings. In order to do so, we try to formulate an objective measure of computational intelligence and study how it presents in human beings, animals and machines. Analogously, we study phenomenal consciousness as a dichotomous variable and how it is distributed in humans, animals and machines. As phenomenal consciousness and computational intelligence are independent, this fact has critical implications for society that we also analyze in this work.}


\section{Introduction}
The concept of consciousness has always been difficult to define, as it is a property that can not be directly measured by any method coming from our current scientific method \cite{searle1995mystery}. In particular, it is an ontologically subjective phenomenon, and a measure in the sense of the scientific method applies to an epistemologically objective phenomenon \cite{timmermans2015can}. However, we can split the consciousness concept into several phenomenons associated with the term to make it easier to work with this phenomenon. Specifically, the philosophy of mind literature splits consciousness into phenomenal consciousness, also referred as vigilance according to neuroscience \cite{dehaene2014consciousness}) and that basically refers to the subjective and private to the observer sensations that human beings feel (also called awareness) and access consciousness, \cite{chalmers2002philosophy}, that refers to the ability to put attention to a particular feeling. Last, another type of consciousness is self-consciousness, that is basically the model of our identity that we build based on our experience and abilities \cite{ravenscroft2005philosophy}. In other words, we differentiate between subjective phenomenona, the relative to the observer experience of something that the observer is paying attention to, and the objective phenomena, that can be simulated in a computer even if it is related to consciousness \cite{merchan2020machine}. For example, we can manipulate colours with computers but we cannot simulate qualia. We can even simulate the bottleneck of consciousness \cite{garrido2022global} but cannot simulate the perceived experience of the result.  

Whereas access and self-consciousness can arguably, and in our opinion, be simulated by a computer successfully, for example with attention mechanisms \cite{vaswani2017attention} and a computational ontology of a person based on historical records \cite{chandrasekaran1999ontologies}, how to implement or simulate phenomenal consciousness, that is going to be the target of this paper, remains a complete mystery. In particular, we can not simulate the experience of the perception of qualia, such as the redness of the red colour. Philosophy of mindexplains this impossibility with the example of Mary, a blind girl that is an expert in vision. Although being an expert, Mary is unable to know what is the difference in the perception of red or black, as this information comes from qualia and is subjective, that is, cannot be represented in an objective manner such as in a book description \cite{ludlow2004there}. This is known as the knowledge argument: and it rests on the idea that someone who has a complete objective knowledge, from the point of view of the scientific method and our epistemological scope, about another conscious being might yet lack knowledge about how it feels to have the experiences of that being \cite{nida2002qualia}. A common belief shared by many people of the computer science community, and concretely in the machine consciousness community \cite{gamez2018human} and even in the philosophy of mind community and more concretely shared by the connectionism community \cite{bechtel1991connectionism}, is that if an artificial general intelligence \cite{goertzel2007artificial} is modelled (supposedly via some meta-learning \cite{vanschoren2019meta} or transfer learning methodology \cite{torrey2010transfer} applied to high capacity deep learning models \cite{lecun2015deep} and huge datasets), due to emergence, phenomenal consciousness may arise. Hence, this group of people believe, assuming the multiple realizability philosophy of mind assumption \cite{heil1999multiple}, that an intelligent enough system is the cause of phenomenal consciousness, that phenomenal consciousness arises as an epiphenomenon \cite{morgado2019consciousness}, or that intelligence is the cause of phenomenal consciousness or viceversa, that is, that they are dependent variables. 

In order to continue analyzing the potential statistical, or even metaphysical, causal relation between phenomenal consciousness and intelligence, it is important to also briefly describe intelligence, which is another controversial concept that has been in focus since the times of ancient philosophers. Coming from the psychology community, and in a broad sense, intelligence is a very general mental capability that, among other things, involves the ability to reason, plan, solve problems, think abstractly, comprehend complex ideas, learn quickly and learn from experience \cite{gottfredson1997mainstream}. If these problems are computational, we can reduce and quantify intelligence as an analytical expression \cite{chollet2019measure}, as we will see in further sections, giving rise to the concept of computational intelligence. From an ontological objective point of view, independent of the observer, the described definition of intelligence shown by a system or living being would be epistemologically subjective, as it is relative to the observer and its knowledge, culture and beliefs. Consequently, we can not directly measure it without observing the external behaviour of the person. As a result, the observer could be really more intelligent that he appears to be according to its behaviour. However, it is the only measure that we can express quantitatively and to study its relation with phenomenal consciousness, although it is based on the behaviour of the subject. As we will further see, as in the case of Asperger and Autism syndrome subjects, the person may be really more intelligent according to Stern's intelligence quotient or some measures of intelligence that we will show than its behaviour shows, which is very problematic. In other words, we argue that computational intelligence would be an ontological objective continuous numerical latent variable whose observation is noisy and obscured by a series of factors such as the personality or mood of the person being measured. As phenomenal consciousness is also an ontological dichotomous property, it only has sense to use this definition of computational intelligence, and not a epistemological subjective definition, to study its independence with phenomenal consciousness.

If these computational problems can be solved by a computer, either traditional or a quantum computer, we find the concept of artificial or machine intelligence \cite{legg2007universal}. Concretely, it is important to remark that the artificial intelligence models implemented in quantum, traditional or other computing paradigms such as biological computation mechanisms, solve the same space of problems, in particular, those solvable by an universal Turing machine. However, artificial intelligence does not involve concepts as understanding \cite{garrido2022artificial}, as understanding requires an entity being aware of the learned concept. Nevertheless, computational intelligence does not require understanding. For example, a model can beat any human at chess, displaying a bigger estimation of computational intelligence than aware beings, while being unaware. It wins chess by having learned complex patterns of previous data with statistical models or performing search algorithms that humans can not perform but it does not understand the meaning of chess as it is not aware of the experience of playing chess. Consequently, it is merely following a sequence of instructions that maximizes the probability of winning chess based on previous data. Hence, we find that computational intelligence, which is a subset of intelligence, is independent in this example from phenomenal consciousness, which is the focus of this paper and we will further continue to provide examples such as this one. General intelligence \cite{wheaton2006towards} involves other types of intelligence such as emotional intelligence \cite{salovey1990emotional} or multiple intelligences \cite{kagan2000multiple}. These intelligences can also be measured in human beings with measures such as the Stern's intelligence quotient \cite{stern1914psychological}, or more sophisticated metrics, and may be simulated in machines if they do not require understanding nor qualia experience. All the other aspects of intelligence coming from the psychology community definition such as learning from experience, thinking abstractly or planning can be simulated with machines, as for example The Generalized Agent (GATO) from DeepMind empirically shows. Citing the paper explicity:

\begin{center}
\textit{"The agent, which we refer to as Gato, works as a multi-modal, multi-task, multi-embodiment generalist policy. The same network with the same weights can play Atari, caption images, chat, stack blocks with a real robot arm and much more, deciding based on its context whether to output text, joint torques, button presses, or other tokens."}
\end{center}

We will provide arguments to theoretically support the claim that computational intelligence is not related to phenomenal consciousness and that machines do not possess phenomenal consciousness. Therefore, as we will illustrate in detail in further sections, we justify and claim that rights must belong to people and ethical decisions must be taken by people, not just by, for example, machine learning systems. Moreover, we must not judge a person for its intelligence but because it is a human being with the potential of being conscious. Not doing it so implies a discrimination of a conscious being because of its measurable computational intelligence.

In this paper, we will provide solid arguments coming from computer science, neuroscience, physics, psychology and philosophy of mind to reject the hypothesis that phenomenal consciousness computational intelligence are dependent variables. We find that the root of this belief is explicitly formulated in the "Analogy" paper written by Bertrand Russell \cite{russell1948analogy}. Hence, all the argumentation that will be shown in this paper tries to show that the Analogy of Russell, that very briefly states that as the observable degree of computational intelligence of an individual increases then it becomes more probable that the individual possesses phenomenal consciousness, is a fallacy.

The organization of the paper is the following.First, we illustrate the Analogy of Russell and formalize it from a Bayesian point of view. Then, we provide some simple counter-examples of it that show empirical evidence of how unlikely is that hypothesis to be true. In an additional section, we study the concept of intelligence and provide a new definition of computational intelligence to more formally reject the mentioned hypothesis. With that definition, modelling computational intelligence with a numerical variable and phenomenal consciousness with a dichotomous variable, we study the potential causal relation of phenomenal consciousness and computational intelligence and theoretically provide an argument that shows how these properties are independent, formally and with an exhaustive list of counter-arguments. Afterwards, we formalize how phenomenal consciousness and computational intelligence are independent and the consequences that this fact have in our society. Finally, we finish the paper with a section of conclusions and further work. 

\section{Russell's analogy of consciousness}
In this section, we will present the analogy postulated bn y Russell about intelligence and consciousness \cite{russell1948analogy}. Broadly speaking, he states that it is highly probable that consciousness is the only cause of the intelligent behaviour that humans exhibit. It does so by supposing that if the behaviour of people is similar to our own, then, by observation, we can establish a causal relation that the other people possess consciousness as we do. Literally, from the Analogy of Russell, we have that:

\begin{center}
\textit{"We are convinced that other people have thoughts and feeling that are qualitatively fairly similar to our own...it is clear that we must appeal to something that may be vaguely called analogy. The behavior of other people is in many ways analogous to our own, and we suppose that is must have analogous causes. What people say is what we should say if we had certain thoughts, and so we infer that they probably have these thoughts... As it is clear to me that the causal laws governing my behavior have to do with thoughts... how do you know that the gramophone does not think? ... it is probably impossible to refute materialism by external observation alone. If we are to believe that there are thoughts and feelings other than our own, that must be in virtue of some inference in which our own thoughts and feelings are relevant... establish a rational connection between belief and data...From subjective observation I know that A, which is a thought or feeling, causes B, which is a bodily act, whatever B is an act of my own body, A is its cause. I now observe an act of the kind B in a body not my own, and I am having no thought or feeling of the kind A. But I still believe on the basis of self-observation, that only A can cause B. I therefore infer that there was an A which caused B, though it was not an A that I could observe."}
\end{center}

Consquently, at least probabilistically, we can infer from the text that behaviour is a cause of consciousness, although we can not observe consciousness. As behaviour is a cause of intelligence, we can infer that intelligence is an effect of consciousness, being a causal relation. However, the previous reasoning is a fallacy, as it deals with the assumption that the intelligent behaviour of any agent shares the same causes as our behaviour, which as he writes in multiple times in the text, is the mind. For instance, the DALLE-2 model generates artistic images but if we assume that the multiple realizability postulate is false, then, it is not conscious, so consciousness is not the cause of its behavior. In other words, it seems that the analogy reasoning involves more correlation that causality. Here the confounder would be that both my behaviour and the behaviour of the DALLE-2 model are both the result of a human intent. From a classical logic point of view, he states that every living being produces intelligent behavior, applying modus ponens. However, applying modus tollens if a being does not exhibit intelligent behavior outside, then it would be no conscious, at least, from a probabilistic point of view, probably. This fact that is a consequence of applying the analogy reasoning has dramatic consequences that will be analyzed in further sections. Moreover, he states that a rational connection between belief, consciousness, and data, behaviour, must be set. Therefore, consciousness would be the cause of intelligence, at least from a computational point of view. Also, we cannot establish a causal relation by performing such an experiment based on analogy alone. In that particular experiment, we observe that a subject is intelligent and we infer that he is conscious. First, we do not have any control group in this experiment, being what is known in science and in a plethora of disciplines such as psychology or econometrics a pre-experiment, an experiment where we are not able to infer a causal relation between the dependent variable, being conscious, and the independent variable, showing an intelligent behaviour.

In addition, we have several other problems, as we will detail in this work. First, dealing with several syndromes and illnessess. Let us examine a person suffering from severe autism, this person may not show an intelligent behaviour, hence, following the argument of Russell, there is, at least, a high likelihood that this person is not conscious. We believe and will add theoretical evidence that this is not true, showing clear empirical cases where a phenomenal conscious human being does not exhibit an intelligent behaviour. Moreover, the same happens with other syndromes such as Down. 

Having shown that there are phenomenally conscious human beings that do not exhibit intelligent behaviour according to several estimates or that its computational intelligence can not be compared to the ones of computers, we provide another counter-example to the analogy, coming from the field of artificial intelligence \cite{russell2002artificial}. In particular, we have seen how, in recent years, due to methodological advantages such as deep learning \cite{lecun2015deep} and the rise of computational power, intelligent systems have surpassed human abilities in a series of complex games. Some examples include AlphaGo winning at the Go game to the world champion \cite{wang2016does}, IBM Watson winning at Jeopardy \cite{ferrucci2012introduction} and discovering new unknown chess strategies with deep learning \cite{mcgrath2021acquisition}. General intelligence is a broad property, in the ontological sense, but we can reduce its meaning and provide a definition for a subset of it. In particular, we can epistemologically measure it as a function of the proportion of the computational problems that a system can solve from the set of all computational problems. Following this lower bound of general intelligence, a system implementing all the known machine learning models and meta models of them able to solve any task with sufficient data will, for sure, outperform the performance of human beings in a broad series of problems and even solve problems that we do not know how to solve, like the protein folding problem \cite{jumper2021highly}.

As a consequence, following the analogy of Russell, it would be highly likely that a system that implements these algorithms would be conscious. It would be even more probable that such a system is conscious that any other human being. However, as we will further show, providing multi-disciplinary arguments, the likelihood that a Turing machine, that is essentially any known software being executed by a computer, is conscious is almost zero. Hence, the evidence given by the data shows that the hypothesis that there is a causal relation between intelligence and consciousness is fallacious. Finally, as an extreme argument, if we measure the computational intelligence of a severe autistic person with respect to a system implementing several methodologies such as AlphaGo, the difference in intelligence, measured for example with the intelligence quotient would be even higher, however, and also with theoretical and empirical evidence coming from the global workspace theory \cite{dehaene2014consciousness}, the severe autistic person is conscious and the system is not. If we follow this argument, we discover that it is more probable that intelligence and consciousness are just independent variables. In this paper, we will slowly give arguments that support this claim.  

Bayesian modelling of the analogy. The correlation between intelligence and phenomenal consciousness is positive and strong, near to $1$. That is, the conditional probability (C) that an entity possesses consciousness given that is intelligent (I) is close to 1 $p(C/I) \approx 1$.  \cite{degroot2012probability}.

Implies a measure of consciousness and intelligence. \cite{searle2008reductionism}

Consciousness cannot be measured with the current scientific method. (Include citations) 

Moreover, can we measure, or even define, intelligence? At least, can we even define different types of intelligence? \cite{wang2009abstract}

\section{Defining intelligence}

Intelligence is a widely known concept that has been assimilated by the computer science community to coin the term artificial intelligence. However, artificial intelligence is a misleading term, as it requires a proper definition of intelligence as a property that can be modelled with a set of numerical variables.

In particular, multiple definitions of intelligence have been proposed by different communities but all of them seem to be a reduction of the general meaning of intelligence. For example, if we include as intelligence the ability to understand and empathize with another person, this ability requires to feel the situation that is having the other person. From a theoretical, relative to the observer and internal point of view, it will not be enough to appear to understand or feel by simulation methods based on quantitative measures, it would need to recieve the qualia of the feeling or the idea being understood.   

Hence, feeling requires awareness, or phenomenal consciousness, of the person that is having a conversation. As a consequence, as we will argue, this ability can not be reduced to a a simple set of numerical variables nor be implemented in a machine.

Consequently, the term artificial intelligence is a reduction of the intelligence of conscious beings and the term intelligence is really hard to define. In this paper we make a review of the different definitions of intelligence, we will technically argue why artificial intelligence is a reduction of biological intelligence and propose a new definition of general intelligence that can not be reduced to a set of numerical variables nor, as we will show, be implemented in machines.

Definitions of intelligence \cite{wang2009abstract}: machine \cite{legg2007universal}, technical or computational (related to artificial intelligence) \cite{chollet2019measure} general \cite{wheaton2006towards} emotional \cite{salovey1990emotional} multiple \cite{kagan2000multiple}, natural \cite{wang2009abstract}. 

\subsection{Artificial intelligence and deep learning}
Artificial intelligence \cite{russell2002artificial} has another controversial definition. Generally, it is the science and engineering of making intelligent machines \cite{mccarthy2007artificial}. But, if we want to define the intelligence of machines, that definition is circular, hence invalid. We prefer to define it as an objective quantitative measure that is determined by the scope of problems that an artificial system is able to solve. 

In the recent years, due to the significant advances of computational power, it has been possible to implement high-capacity machine-learning models \cite{murphy2012machine} like deep neural networks, what is usually referred as deep learning \cite{lecun2015deep}. As we have illustrated in the introduction, these models, whose capacity includes having more than 500 billions of parameters \cite{ren2021zero}, are able to solve complex problems like the protein folding problem \cite{jumper2021highly}, go and chess \cite{wang2016does}, write philosophy articles in a newspaper mocking the type of writings that were usually only attributed to human beings \cite{floridi2020gpt} mastering natural language processing and common sense tasks and generating art \cite{wang2017multimodal}. In essence, deep learning methodologies are able to fitting complex probability distributions being able to generalize their behaviour to tasks that we only supposed to be solved by humans, making their behaviour indistinguishable from the one of humans \cite{elkins2020c}. 

However, deep neural networks are software programs that are executed using computer hardware in a CPU (Central Processing Unit), GPU (Graphical Processing Unit) or TPU (Tensorial Processing Unit). Concretely, these hardware units are a part of a Von Neumann architecture, that is essentially a Turing machine, making deep neural networks algorithms that can be executed by a Von Neumann architecture, hence a Turing machine. Consequently, as we will further provide arguments for this claim, they lack awareness or phenomenal consciousness. As a result, they are unable to understand nor experience the scope of problems that they are solving and merely solve computations involving pattern recognition, independently of their complexity. Hence, artificial intelligence systems only posses computational intelligence \cite{legg2007universal} \cite{chollet2019measure}, lacking understanding as it requires the qualia of the problem being solved. However, a virtue of computational intelligence is that it can be quantified, as it solves objective problems belonging to the set of all possible computational problems. In contrast, general human intelligence, as we will further see, is a subjective and relative to the observer measure, requiring the qualia generated by understanding, feelings or empathy and hence being impossible to fully quantify without incurring in a reductionist measure. 

There are several propositions to quantify the computational intelligence that a system or an entity possess. Let $\pi$ be an entity, for example a human being, that in every instant $t$ is able to perform a set of actions $A$ to solve a given problem. An intelligent agent $\pi$ would decide, for every instant $t$, the optimum action $a^\star \in A$ to solve the problem. The branch of computer science that studies how to train intelligent agents in this framework is called reinforcement learning \cite{sutton2018reinforcement}, and can be directly extrapolated to the reality. For example, if we want to say the optimum phrase to win a negotiation, in every instant $t$ we receive the sentence of the person that we are negotiating with, its word frequency, mood state and more information and as a function of all that information we choose to answer a certain phrase in a particular mood state. As we can see, the reinforcement learning can be applied to a plethora of computational intelligence problems. In fact, reinforcement learning systems are implemented in robots for planning. Dealing with these systems, that can perfectly be humans, the universal intelligence function $\Upsilon$ of a data structure resembling an agent $\pi$ is given by the following measure \cite{legg2013approximation}:

\begin{equation}
    \Upsilon(\pi) = \sum_{\mu \in E} 2^{-K(\mu)}V_{\mu}^\pi\,.
\end{equation}

where $\mu$ is a data structure representing an environment from the set $E$ of all computable reward bounded environments, $K(\cdot)$ is the Kolmogorov complexity, and $V_\mu^\pi := \mathbb{E}(\sum_{i=1}^\infty R_i)$ is the expected sum of future rewards $R_i$ when agent $\pi$ interacts with environment $\mu$. That is, the previous expression is a weighted average of how many problems $\mu \in E$ does an agent $\pi$ solve weighted by their difficulty $2^{-K(\mu)}V_{\mu}^\pi$ and the capabilities of the agent. In particular, this is the reason why $V_{\mu}^\pi$ includes $\pi$. Several things are interesting dealing with this expression. First, the set of all computable reward bounded environments, \textit{i.e.}, generalizing this set would be the set of all computational problems, is countably infinite, having as an axiom that it exists a particular simple problem that can not be decomposed in more single parts. Hence, the intelligence $\Upsilon$ of an agent is not upper bounded. If we transform the set $E$ to a set where the area of a problem $\mu$ is given as a function of its difficulty $S(\mu)$, being more area given to more difficulty with respect to a particular agent $\pi$, the previous measure can be transformed in this abstract, general measure: 

\begin{equation}
    \Upsilon(\pi) = \int_E \delta(\mu|\pi) S(\mu) d\mu\,.
\end{equation}

where $S(\mu)$ is a oracle function that gives the objective area of a problem $\mu$ and $\delta(\mu|\pi)$ is a delta function representing whether the particular problem is solved or not by the agent $\pi$. Recall that the delta function outputs $1$ if the problem is solved and $0$ otherwise. As the set is potentially countably infinite, a problem can be decomposed according to the progress on it in different problems until a simple base problem, each one with different area to measure the progress of an agent in the progress of a particular problem. Interestingly, the integral over the set $E$ gives the area of computational problems being solved, and this area is infinite. Moreover, an oracle giving the particular objective unbiased measure of difficulty for every problem would be needed. Depending on the features of the system, a problem may be more difficult than other, specially for non-computable problems requiring qualia to be solved. These objections make such a measure impossible to be unbiasedly implemented in practice, but may be a lower bound of the computational intelligence of a system, animal or human being.    

Another example of an intelligence measure of a system represented by $IS$ for a scope of tasks sampled from $P_{scope}$ is now described. We have generalized from the measure proposed by Chollet, taking into account not only a scope of particular task that are numerable into a set but all the possible tasks that can be done in our universe, which is potentially infinite and the one that we believe that should be taken into account. Recall that we want to provide an ontologically objective general measure of computational intelligence \cite{chollet2019measure}, as we want to study its independence with an ontologically objective dichotomous property, that is whether an agent is aware of its phenomenal consciousness. Consequently, any measure that excludes a single property or is noisy or biased, such as the intelligence quotient, can not be compared with phenomenal consciousness without being also the results biased or noisy. Summarizing the main components of the expression, let $P_{IS,T} + E_{IS,T,C}$ (priors plus experience) represent the total exposure of the system to information about the problem, including the information it starts with at the beginning of training represented by $C$, the curriculum. Let $\omega_T \cdot \theta_T$ be the subjective value we place on achieving sufficient skill at $T$ and let $GD$ be the generalization difficulty for agent $IS$ of solving task $T$ given its curriculum or specific properties of the agent $C$: 

\begin{equation}
    I_{IS, P_{scope}}^{\theta_T} = \mathbb{E}_{P_{scope}}[\omega_T \cdot \theta_T \sum_{C \in Cur_{T}^{\theta_T}} [P_C \cdot \frac{GD_{IS,T,C}^{\theta_T}}{P_{IS,T}^{\theta_T} + E_{IS,T,C}^{\theta_T}}]].
\end{equation}

The formula is basically a generalization of $\Upsilon$ that takes into account the previous knowledge, modelled by the curriculum and the priors, to solve a particular task $T$. The difficulty of the task is now modelled by the generalization difficulty and solving a potentially infinite scope is given the computing the expectation over $P_{scope}$. However, although this measure takes into account whether an entity is able to generalize from prior knowledge as a measure of intelligence, we find the same problems than in the previous measure.  

In both measures of intelligence, as the set of potential problems is potentially infinite and not-numerable, any entity would really have a measure of general intelligence of approximately 0, as it would fail to solve a potentially infinite set of problems. Moreover, both measures require having an oracle to determine the difficulty of the task. Consequently, they would both be biased although an objective oracle was able to provide this quantity. 

Any measure of intelligence giving any other score rather than zero, although practical, would be just a lower bound of the true intelligence of the entity, better approximated with these measures than with the intelligence quotient measure. Hence, it can be useful for health situations but never to classify an individual as more intelligent than another individual or, as we will further see, to say that a being is susceptible of having more or less likelihood of having phenomenal consciousness as a result of scoring more degree of intelligence according to a measure, as it is only a lower bound on the true intelligence, that cannot be measured in practice as we can only approximate it based on a subset of problems and by the external behaviour of the subject as in the case of the intelligence quotient. However, recall that it is critical to provide an abstract definition of computational intelligence because of two main reasons: first, in order to study its independence with an ontological property such as being aware of phenomenal consciousness, it needs to possess the same properties as phenomenal consciousness, that is, being ontological, general and not biased. As an analogy, in statistical terms, it needs to be defined as the parameter, from a frequentist point of view. Second, it can be useful to provide such a definition of intelligence to shed light to the psychology community to provide less biased estimators to it. Once again, this definition corresponds to the parameter and measures such as the intelligence quotient correspond to the estimator.     

\subsection{Intelligence quotient and similar approaches} 
Human intelligence includes a series of skills that are able to solve different types of problems. The set of problems that human intelligence can solve intersects with the set of computational problems but is not contained on it.

Some examples of these kind of problems include discriminating which is the most beautiful color for a particular observer in terms of our perception of the colours, which is the best action that we should do in a complex personal conflict involving human relationships, how do a person change its state of mood or which is the true notion of a metaphysical phenomenon. The common feature of all these problems is that they involve qualia, an information about our universe that Turing machines lack. In particular, we consider qualia as semantic information, in the sense that the observer perceives the quale of a color in a particular way, the redness of red, and not in another one. Consequently, this perception can be considered a property that may be codified and that is actually transmited to the observer by the brain. Although this information is subjective and relative to the observer, it is still information that can be represented in a qualia space such as in the integrated information theory and is transmitted to the phenomenal consciousness observer.

Consequently, from our point of view, we can only measure the intelligence that a human being shows externally and that is associated to these problems in terms of correlations, that are a reduction of its true scope but are the only way of being objective. Since ancient times, human intelligence has been measured with features. For example, in the ancient Greece memory was very valuable, then the Roman society paid special attention to rhetoric. In the past century, abstract reasoning was very appreciated and become a critical feature of Stern's intelligence quotient \cite{stern1914psychological}. Stern's intelligence quotient assigns a mental age to a person based on her performance on a series of tests including reasoning, logic, language and more. In particular, he divides the scored mental age with the chronological age to obtain a simple ratio. 

However, several features that are independent of intelligence may affect Stern's measure. For example, the subject can be in a sad mood, be an introvert or have some special condition as autism. Due to these conditions, the intelligence shown externally by the subject does not correspond to its true intelligence, in other words, the true human intelligence would be a latent variable contaminated by noise or any approach that measures human intelligence as the Stern's intelligence quotient is an approximation to the underlying intelligence of the subject.

Moreover, as Stern's test and similar ones include only a subset of all the subjective problems that a human being is able to solve, the intelligence measured by these tests is a lower bound of the true intelligence of the human being. Consequently, we believe that these approximations are very naive, poor, unreliable, culturally biased and noisy. From a statistical point of view, the intelligence quotient would be a poor estimator of human intelligence: biased because it does not test all the areas of intelligence and it is influenced by western culture and with high variance as its measurement contains noise because individuals may be nervous, be shy, have a special condition or simply do not wish to score high.   

Hence, as we can only obtain a measure of intelligence via a test analogous as the one of Stern, as the quality of the approximation is poor, the value of this random variable cannot be used in a causal relation with the value of the phenomenal consciousness dichotomous variable. Recall that these tests are only able to reduce the true underlying intelligence of a human being, or even a system, as an approximate lower bound. Consequently, this quantity can not be established as the cause nor the effect of phenomenal consciousness. 

We can illustrate several examples of this statement. First, we would assume that Russell is correct and consciousness is an effect, or an epiphenomenon, of intelligence. Firstly, a comatose person is, according to neuroscience, phenomenally conscious \cite{bayne2020there} but would score a 0 according to Stern's test or similar ones. Hence, it would not be probable that this person is conscious if there is a causal relation between consciousness and intelligence and by analogy with respect to other human beings as Russell describes in his text. However, it can be \cite{bayne2020there}. Another example includes a natural language generative transformer like GPT-3. This algorithm is very close to passing the Turing test \cite{elkins2020c} and performs greatly in intelligence quotient tests, however, as we will see in the further section, the system does not possess awareness. Finally, a Down-syndrome person would score less points in average than a neurotypical person but both possess consciousness. These three examples show how, at least computational intelligence and phenomenal consciousness are not related, so the Russell's analogy is false. An even more convincing case than the rest is this one: In the science-fiction book The Three Body Problem \cite{liu2014three} an enormous plethora of people was displayed in a planet like a CPU. Each person acts as a transistor, creating a huge Von Neumann architecture. Most critically, observe that there does not exist a physical connection between the people acting as transistors. Consequently, according to consciousness theories such as information integration theory, that requires physical connections \cite{tononi2008consciousness}, or the Pribram-Bohm holoflux theory of consciousness \cite{joye2016pribram}, this people CPU would be non-conscious as a whole. However, it is able to solve the same problems that a high capacity deep learning model is able to solve, as the people CPU can execute a program that implements the deep learning model. This is the most obvious case where we can see that any algorithm, independently on the degree of intelligence that we can measure with respect to its behaviour, does not have phenomenal consciousness and that phenomenal consciousness is independent from intelligence. In the following section we will argue how non-computational intelligence may be correlated by consciousness, but that it remains a mystery and we can not say objectively if they are dependent or not. 

\section{Intelligence is not a measure of consciousness}
If we accept, as an absurd, that are dependent, we find some problems. Use in all the section Bayes theorem to model the two hypotheses, entities having or not phenomenal consciousness.

\subsection{Machine consciousness}
The computer science community that studies the potential for consciousness in machines is called machine consciousness \cite{gamez2018human}. In particular, the machine consciousness community, inherits the assumptions of functionalism, like multiple realizability, and connectionism to especulate that systems or robots may develop qualia through the implementation of expert systems, machine learning models, hybrid methodologies or other variants of information processing systems that, in any case, they can be emulated using Turing machines \cite{gamez2018human}. However, as we have illustrated in the previous section with the people CPU, the computational intelligence shown by algorithms, independently of its complexity, is not the cause of phenomenal consciousness. Moreover, as we will illustrate in the following section, there are more philosophical arguments that provide evidence on the highly remote hypothesis that computational intelligence is the cause of phenomenal consciousness.

\subsection{Strong artificial intelligence counter-arguments}
Our argumentation depends on the assumption that artificial intelligence systems, like high-capacity deep learning models, are not aware of themselves. As currently an ensemble of systems would have greater computational intelligence than human beings and they do not have phenomenal consciousness, this example is a great counter-argument to the hypothesis that phenomenal consciousness is an epiphenomenon of computational intelligence or that they are simply dependent variables. Hence, in this section, we will describe the main counter arguments to the strong artificial intelligence hypothesis, i.e., the one saying that complex machines implementing high capacity models and reasoning systems may arise consciousness by emergence.

The nobel laureate Roger Penrose, defending its controversial Orch-Or theory that states that phenomenal consciousness arises at the quantum level inside neurons  \cite{penrose1991emperor}, gives a plethora of strong artificial intelligence counter-arguments in its books. 

First, we find the famous Searle chinese room \cite{searle2009chinese}. This experiment basically denotes the difference between pattern recognition and understanding. Generalizing the argument, suppose that an entity is hidden in a room where a text written in an unknown language is read by the entity. The entity has a dictionary, or a mapping function, with the correct answers to the questions written in that language. From the point of view of an observer located outside of the room, the entity that resides in the room appears to understand the language, however, the entity does not understand nothing, as it lacks an understanding of the language.

From our point of view, understanding a language requires an additional mapping that the observer that lies inside of the room lacks. A mapping of every word of the language and the qualia that the words refer to. Qualia is necessary for understanding, and phenomenal consciousness is necessary for qualia. Hence, as the machines lack phenomenal consciousness, they are unable to understand a language and consequently all that they do perform is pattern recognition, in other words, solving complex correlations creating a function whose input is a sentence of a language and its output is another sentence of that language.

Recall from previous sections, where we provide the example of the people CPU that appears in the science fiction book The Three Body Problem \cite{liu2014three} that was able to perform complex computations and run algorithms to predict a planetary disaster without using computers, that performing complex pattern recognition tasks due to the information processing done by high capacity deep learning or other statistical models is not enough to arise phenomenal consciousness by emergence. Concretely, not only in science fiction we have found an example of a person CPU, in a real experiment available on Youtube \cite{stilwell2018brain} and that has been implemented in a code that is available on Github, we have found how people were organized smartly in a field emulating a brain to perform an algorithmic task. If that experiment had more people available, they could solve any kind of problem that a Turing machine is able to solve. In other words, the Stilwell brain is also a Turing machine, as quantum or classical computers, that does not posses phenomenal consciousness. If an external observer does not know whether the Stilwell brain is a code, as the one hosted on Github, or people being organized in a smart way, it could argue that is intelligence, hence following Russell's analogy potentially phenomenally conscious, however, according also, without loss of generality, to the integrated information theory of consciousness and the Pribram-Bohm holoflux theory of consciousness, the Stilwell brain or any other brain created by independent entities is an excellent example that shows how phenomenal consciousness and intelligence are independent.

It is also important to consider that all the algorithms that can be executed in a computer can be solved by Turing machines \cite{hopcroft1984turing}. Quantum computers are not an exception, both classical computers and quantum computers are universal Turing machines and, hence, solve the same kind of problems only with different computational complexity \cite{deutsch1985quantum}. Nevertheless, humans are able to feel, that requires being able to perceive the qualia of the feeling and, we have said before, having the phenomenal consciousness required to feel, phenomenon that is not able for a Turing machine and that we can not measure objectively \cite{penrose1994shadows}. If quantum computers are not able to possess the characteristics and abilities of phenomenal consciousness, hence, the idea of the brain being a quantum computer or arising phenomenal consciousness by means of a quantum-like procedure is, at least, a reductionist one, as, in principle, phenomenal consciousness is independent of this procedure. 

Finally, because of the qualia that we experience, we can gain an intuition about problems that do not have an algorithmic solutions. It is specially relevant that this intuition, the experience of being able to understand these problems, can not be sensed by a computer, as it can not perceive qualia. Some examples of these problems are the following ones. First, the halting problem, that is, being able to determine, from a random computer program description and an input, whether the program will finish executing the problem, or continue to run forever \cite{lucas2021origins}. Second, Hilbert's tenth problem dealing with Diophantine equations, that are equations involving only sums, products, and powers in which all the constants are integers and the only solutions of interest are integers. In particular, the Hilbert's problem, proved to be undecidable, is described as being able to find an algorithm that decides whether a random Diophantine equation has an integral solution \cite{matilasevich1993hilbert}.  

\subsection{Disability and comatose states}

Having seen that solving a wide variety of tasks inside the set of all computational problems, even more than the ones of human beings, does not require that the computationally intelligent system has phenomenal consciousness  we will now study another causal relation that, according to Russell's analogy, would incur in a low likelihood for the entity to be conscious. It is the one dealing with people suffering different syndromes as Down \cite{epstein1989down} or severe Autism \cite{lord2000autism} that would make them score low in Stern's intelligence quotient test and that, however, are phenomenally conscious. Once again, we find another counter-example to the Russell's analogy dealing computational intelligence. Moreover, autism is a curious case, as it is highly correlated with special abilities such as the one of Daniel Tammett \cite{tammet2007born} that is able to memorize and say the first 22514 digits of number pi in only 5 hours, or speak 11 languages. According to intelligence quotient tests the abilities of Tammett would not be quantified, hence being the tests a lower bound on intelligence and being intelligence a variable that is not correlated with phenomenal consciousness, as other social abilities are hard for ASD people like Tammett. 

The extreme case would be the one dealing with a comatose person. Concretely, there is empirical evidence coming from neuroscience that shows how comatose people have neural correlates of consciousness, what has been called as islands of consciousness \cite{bayne2020there}, conscious states that are neither shaped by sensory input nor able to be expressed by motor output. Technically, people suffering a comatose state would be phenomenally conscious but unable to perform any kind of movement nor reaction to any external stimuli. Consequently, the score that they would perform in any kind of test similar to the one of Stern's intelligence quotient would be 0. Nevertheless, the neuroscience community shows evidence to support the claim that they are phenomenally conscious or just aware but unable to report any stimuli. 

\subsection{Consciousness in the animal kingdom}
Neurobiology gives us evidence that animal brains share features with our brains dealing with the neural correlates of consciousness \cite{griffin2004new}. Concretely, this evidence does not only reduce to the most similar animals to us, like primates, but these neural correlates are found, up to some degree, in other mammals, birds, and at least some cephalopod molluscs, like octopuses, squid or cuttlefish \cite{birch2020dimensions}. 

Following Rusell analogy, concerning intelligence, we can say that the neuroscience community gives evidence to suggest that animals, young infants and adult humans possess a biologically determined, domain-specific representation of number and of elementary arithmetic operations \cite{dehaene1998abstract}. However, we have seen that computational intelligence seems to be independent with phenomenal consciousness. Nonetheless, phenomenal consciousness is a prerequisite to experiment the qualia of subjective phenomena such as being aware of feelings. Precisely, concerning the qualia of feelings, neurobiological evidence shows how animal brains perform similarly to us dealing with the elaboration of the primary emotions, which include the foraging-expectancy system, the anger-rage system, the fear-anxiety system, the separation-distress-panic system and social-play circuitry \cite{panksepp1989neurobiology}. Consequently, it seems very plausible that, although animals would score very badly in an Stern's like intelligence quotient test, they may have phenomenal consciousness, as their neural correlates of consciousness, i.e., their emotion processing, seem to be similar to ours. 

Ironically, as we have said, Stern's intelligence quotient is culturally biased, but according to our proposed general intelligence metric, the size of the computational set of problems is infinite. Hence, if an animal has, for example, better memory that us, in average, we could argument that a lower bound intelligence quotient biased to memory would make them score better score than us. Dealing with different aspects of intelligence, animals score better than us, in average. Some examples are birds in spatial memory \cite{payne2021neural}, dogs in smell \cite{horowitz2010inside}, ants in visual \cite{graham2017vision} and even bats in abities that are unique to them \cite{nagel1974like} and that human beings would score a zero. Consequently, following Russell's analogy, the likelihood of them having phenomenal consciousness would be greater than us. However, although it is very probable that they are phenomenally conscious due to neurobiological evidence based on the neural correlates of consciousness \cite{birch2020dimensions}, it is not as evident as in the case of human beings. Hence, again, Russell's analogy fails in this case.

\section{Phenomenal consciousness is independent of computational intelligence}
We will now formalize Russell's analogy from a Bayesian point of view. The latent, unobservable measure would be whether an entity possess phenomenal consciousness or not. We assume here, as we isolate the observer of phenomenal consciousness, in the sense of the defined term awareness by Dehaene, from all the different features of consciousness such as access consciousness, that phenomenal consciousness is a dichotomous variable $C$. Recall that phenomenal consciousness is not being aware of more or less phenomena, as the complexity of the integrated information theory qualia space $\phi$ can model. Phenomenal consciousness, from our definition, is being an observer of the qualia space generated by a living being. Consequently, you can only be aware of the qualia space, an observer of the qualia space, or not. Hence, following our assumptions that phenomenal consciousness is not an epiphenomenon or intrinsically related to the qualia space but a property of beings to be aware of their qualia space, we define phenomenal consciousness as a dichotomous variable of perceive or not the qualia space that a being generates.

Let $I$ be the computational intelligence of an entity as have defined it in previous sections, denoted by the continuous numerical variable $I$. A subject $S$ may possess or not phenomenal consciousness, but with the current state of science, we are only able to determine whether it is conscious by looking at the neural correlates of consciousness. If the system does not have a biological brain nor nervous system, science is unable to provide any clue about the consciousness of $S$. Then $p(C|S)$ would be the conditional probability that a subject $S$ has phenomenal consciousness such that $p(C=1|S)+P(C=0|S)=1$ and $p(I|S)$ is the conditional probability of the computational intelligence of the subject. Concretely, an intelligence quotient test would not determine the intelligence of $S$ as a point estimation but the only thing that it would do is to reduce the entropy of the $p(I|S)$ distribution. 

In order to carry out this analysis we use some concepts from probability theory that we now review. The first one is the amount of information needed to encode a probability distribution, also known as entropy. The entropy $H(\cdot)$ can be viewed as a measure of information for a probability distribution $\mathcal{P}$ associated with a random variable $X$. That is, its self-information. It can be used as a measure of uncertainty of a random variable $X$. When the random variable is continuous, we refer to the entropy as differential entropy. The entropy of an uni-dimensional continuous random variable $X$ with a probability density function $p(x)$, or differential entropy $H[p(X)]$, is given by the following expression:
\begin{equation}
H[p(X)] = - \int_S p(x) \log p(x) dx\,.
\end{equation}
Where $S$ is the support of the random variable $X$, that is, the space where $p(x)$ is defined. The entropy $H(\cdot)$ is useful to model the following relation: If we have a random variable $X$ with high entropy $H(\cdot)$, that means that we have low information about the values that it may take. On the other hand, if we consider a random variable $X$ with low entropy $H(\cdot)$, it is a sign that we have high information about the potential values that the variable $X$ can take. In other words, higher knowledge of a random variable implies lower entropy and viceversa. Another interesting concept regarding information theory, that we use in this work, is the mutual information $I(X;Y)$ of two random variables $X$ and $Y$. Mutual information is defined as the amount of information that a random variable $X$ contains about another random variable $Y$. It is the reduction in the uncertainty of one random variable $X$ due to the knowledge of the other. Mutual information is a symmetric function. Consider two random variables $X$ and $Y$ with a joint probability density function $p(x, y)$ and marginal probability density functions $p(x)$ and $p(y)$. The mutual information $I(X;Y)$ is the relative entropy between the joint distribution $p(x, y)$ and the marginal distributions $p(x)$ and $p(y)$:
\begin{equation}
I(X;Y) = \sum_x \sum_y p(x,y) \log \frac{p(x,y)}{p(x)p(y)}\,.
\end{equation}
Concretely, we define as information gain the amount of information that we gain for a certain random variable knowing the value of the other one.

According to Russell, we know that human beings are likely to be are conscious, so we denote the being a human being as the dichotomous random variable $B$. Then, $p(C=1|B=1)=1$ independently on the degree on intelligence. More technically, the information gain of the intelligence degree $I$ over consciousness given that the entity is a human being is $0$.

\begin{equation}
    IG(C,I|B=1) = 0.
\end{equation}

In other words, the entropy $H(\cdot)$ of the conditional probability distribution of consciousness being also conditioned to the degree of computational intelligence of subject $S$, which is also a random variable as we do not have direct access to it, is the same one. Then, in our case we can illustrate that the entropy on the consciousness random variable for humans $H(C|B=1)$ is equal to the conditional entropy on the consciousness for a certain computational intelligence level $I$. 

\begin{equation}
H(C|B=1,I) = H(C|B=1).    
\end{equation}

As $p(C=1|B=1)=1$, there is not need to show that $H(I|B=1,C=1) = H(I|B=1)$, as it is obvious. Hence, we have formally shown how, for the case of human beings, the computational intelligence degree is independent from the phenomenal consciousness variable. However, until now we have only performed the analysis of computational intelligence and phenomenal consciousness in the case that the subject is a human being. Nevertheless, important implications of this analysis need to be taken into account. For example, we now know that a low measure of computational intelligence according to the intelligent quotient of Stern does not condition the subject from being conscious. Let $p(I)<<$ denote a probability distribution over the computational intelligence for a subject $S$ having its density concentrated over a low value. Concretely, we know that $p(C=1|p(I)<<)=1$. We put here $p(I)$ and not $I=k$ being $k$ a real number as we have denote that current measures of intelligence are a noisy lower bound over the true value of intelligence of subject $S$, that is a random variable. Importantly, we now know with complete certainty that, in the case of disabilities or certain comatose states, a subject has phenomenal consciousness.  

Next, we analyze and compare the probability distributions $p(C|I)$ and $p(C)$. Science gives us evidence that if the entity shares features with the human being biologically talking, concretely the neural correlates of consciousness, the subject may be conscious. We denote with $N \in [0,1]$ a continuous numerical variable that represents the degree of biological similarity of the brain of the subject with the brain of the human being. Concretely, current AI systems, denoted with the dichotomous variable $A$, have $A=0$, as deep neural networks or meta-learning methodologies are just sequences of instructions sequentially computable by Turing machines as we have shown before, although the name may be misleading. 

We found a real analogy with $P(C)$ and $P(C|N)$. Concretely, these variables are, according to evidence found in neurobiology, linearly correlated, i.e., $r(C,N) \approx 1$ being $r$ the correlation coefficient. However, a bird, elephant, dolphin, monkey or cephalopod, for example, may score a low computational value $p(I)<<$. However, and again, we find that conditioning the variable $p(I)<<$ to the conditional distribution $P(C|N)$ does not change the entropy of the distribution:

\begin{equation}
    H(C|N,p(I)<<) = H(C|N).
\end{equation}

Finally, we use the example of a meta-learning system to show how the degree of computational intelligence is not correlated to phenomenal consciousness. Concretely, a meta-learning system with $N=0$ has the biggest computational intelligence known as it is able to solve a potentially infinite set of computational problems that humans or animals are not able to solve up-to-date as we have seen in previous sections. We denote that such a system has a computational intelligence probability distribution $p(I)>>$. However, we know that: $p(C=0|N=0)=1$, independently on its degree of computational intelligence. In other words, if we condition the probability to $p(I)>>$, for all the set of artificial intelligence systems, we have that $p(C=0|N=0)=p(C=0|N=0,p(I)>>)=1$. Hence, the degree of intelligence does not generate phenomenal consciousness as an epiphenomenon or by emergence. Concretely, it is the anatomy of the biological brain, or even less probably the nervous system or body, where supposedly we find, at least, neural correlates of consciousness. Given all the information and evidence that we have provided, we could formalize that the information gain of the computational intelligence random variable given that we know the phenomenal consciousness variable if we marginalize the kind of entity that may have phenomenal consciousness is 0, \textit{i.e.}, they are independent random variables independently on the intelligence degree.

\begin{equation}
    IG(C,I) = 0.
\end{equation}

From a Bayesian point of view, this information could be formalized as follows. Concerning artificial intelligence systems, let $p(C=1|I,N=0)$ be an a priori distribution representing the probability of the system being conscious, our previous beliefs coming from the Russell's analogy. Following this analogy, this probability was high as the system is intelligent and the complementary probability, $p(C=0|I,N=0)$, is low. We have provided empirical and theoretical evidence showing that this is not true that we formalize in the likelihood $p(E|C=1, I, N=0)$, being $E$ the evidence that we have illustrated in previous sections. Let $p(E)$ be the marginal likelihood representing the probability of our evidence being true, which is high due to the fact that it comes from highly cited papers of various research communities like neurobiology, psychiatry or philosophy of mind. Lastly, let $p(C=1/E,I,N=0)$ be our posterior beliefs of the hypothesis that artificial intelligence systems are conscious. As the probability rectifier coefficient is very low, that is $p(E|C=1, I, N=0)/p(E)$, despite having an a priori belief supporting the hypothesis of conscious artificial intelligence systems, now the posterior belief clearly shows that $p(C=0/E,I,N=0)>p(C=1/E,I,N=0)$ significantly. Mainly because computational intelligence is independent from phenomenal consciousness.

\section{Repercusions in society}
Given that deep learning and related models, which are a simplification of reality, do not have phenomenal consciousness, that is, they are not aware of themselves, they do not have an identity. Consequently, as they do not have an identity, they can not have the intention nor perceive the idea of inventing nothing, as they are not aware of anything. As a result, a machine can not have the intellectual property of an invention, since it is only the tool of the inventor, as aware of the invention as a pencil or wrench can be. Hence, we can not attribute an invention or discovery to them, as they are not aware of the action of inventing nor discovering. For example, suppose that an economist discovers, using machine learning models or econometrics, that a certain consuming behaviour in the population is correlated with their income. This discovery cannot be attributed to the model, but to the scientist that has formulated the research question and formulated the necessary methodology to obtain significant empirical evidence to support its claim. Similarly, if an architect designs a house and draws it in a paper, we can not attribute the creation of the house to the pencil but to the architect. Consequently, if a deep learning system discovers the solution to the protein folding problem \cite{jumper2021highly} its discovery can not be attributed to the deep learning system but to the scientists team that configured the deep learning system and gathered the data used to fit it.

Hence, we can not attribute a patent to an artificial system, independently on its degree of computational intelligence. Basically because a prerequisite for an entity to have a patent would be that the entity is aware of itself. And, as we have seen in previous sections, artificial intelligence current systems are not aware of itself. Consequently, patents would belong to the team of scientists that build the model or configured the artificial intelligence system.

Artificial intelligence models are a simplification of reality. "All models are wrong, but some of them are useful" \cite{box1979all}. Concretely, they are useful for an aware human being to help him to decide what to do in a complex decision such as a clinical decision, a business problem or discriminating which physical hypothesis is true within a set of plausible ones. We believe that a decision that is potentially harmful for an aware entity, being human or animal, must be at the end executed by an aware entity that is able to comprehend the decision that is being taken. That is, the entity is required to have phenomenal consciousness and a computational intelligence level enough to comprehend the reasoning of the decision. The only entity able to be the responsible for such a decision is the human being. However, the human being can use as a tool a computationally more intelligent system that it for advice. Concretely, explainable machine learning and artificial intelligence \cite{dovsilovic2018explainable} are models and algorithms that are able to justify its decision using a logic that is understandable for humans and would be the best methodology for this scenario. 

Finally, we also consider than in an ethical situation where the integrity of entities is in danger, and some of them can survive and others not, an artificial system can never be priorized over an aware being. Our axiom for the decision is that we should always priorize the survival of the entity that is aware of itself, since phenomenal consciousness cannot be replicated artificially but any kind of artificial device can be replicated. In other words, the value of phenomenal consciousness is much higher than any artificial system. As a corollary, for example, if an artificial intelligence system can be damaged versus an animal or human being damaged, the artificial intelligence system would necessarily be, independently of its complexity, the one damaged. Most critically, our main conclusion is that the priority would be to save the aware entity, and as the artificial intelligence system is not aware of itself, the integrity of the aware entity must be prioritized independently on its computational intelligence degree because its value is higher than any artificial system as, independently on its cost, we can not replicate the exact aware system but we can replicate any artificial machine created by humans. This is a corollary of our assumed axiom, that is priorizing potentially aware entities to non-conscious entities and that conscious beings are not replicable by humans. 

\section{Conclusions and further work}
Phenomenal consciousness is defined as the awareness of an individual to internal and external estimuli, of the information processed by the brain, in the form of qualia. In this work, we have analyzed and show how the Russell's analogy of consciousness, that basically states that awareness and intelligence are correlated with high probability, is a fallacy at least for computational intelligence. In order to do so, first, we defined what is phenomenal consciousness and give an objective measure of computational intelligence. Then, we provided a set of counter-arguments to Russell's analogy with evidences coming from neurobiology, psychiatry or philosophy of mind, where we can see how phenomenal consciousness and computational intelligence are independent. Consequently, we include a formalism with probability and information theory to represent this independence. Finally, we conclude with the social impacts of this fact, mainly that aware beings must be prioritized from non-conscious machines independently of the degree of intelligence, that as machines do not have identity they can not possess patents and that they should not be responsible of decisions that could harm an aware being, mainly because as they do are not aware they can not be responsible of a decision.    

\bibliographystyle{acm}
\bibliography{notes}

\end{document}